\documentclass{article}

\usepackage[preprint]{neurips_2026}

\usepackage[utf8]{inputenc}
\usepackage[T1]{fontenc}
\usepackage{hyperref}
\usepackage{url}
\usepackage{booktabs}
\usepackage{amsfonts}
\usepackage{amsmath}
\usepackage{amssymb}
\usepackage{nicefrac}
\usepackage{microtype}
\usepackage{xcolor}
\usepackage{graphicx}
\usepackage{multirow}
\usepackage{enumitem}
\usepackage{subcaption}
\usepackage{listings}
\usepackage{placeins}
\usepackage{float}

\newcommand{\ours}{\textsc{SOB}}
\newcommand{\numtext}{5{,}000}
\newcommand{\numtest}{5{,}000} 

\title{The Structured Output Benchmark: A Multi-Source Benchmark for Evaluating Structured Output Quality in Large Language Models}

\author{
  Abhinav Kumar Singh\textsuperscript{1} \quad
  Harsha Vardhan Khurdula\textsuperscript{2} \quad
  Yoeven D Khemlani\textsuperscript{2} \quad
  Vineet Agarwal\textsuperscript{3} \\[0.3em]
  JigsawStack, Inc. \\
  \textsuperscript{1}New Delhi, India \quad
  \textsuperscript{2}San Francisco, CA, USA \quad
  \textsuperscript{3}Durgapur, WB, India \\
  \texttt{\{abhinav, harsha, yoeven, vineet\}@interfaze.ai}
}

\begin{document}

\maketitle

\begin{abstract}
Large Language Models are increasingly being deployed to extract structured data from unstructured and semi-structured sources: parsing invoices, medical records, and converting PDF documents to database entries. Yet existing benchmarks for structured output generation either focus on schema compliance alone, or evaluate value correctness within a single source domain. We introduce \ours{} (The Structured Output Benchmark), a multi-source benchmark spanning three source modalities: native text, images, and audio conversations. All models receive a text-normalized representation of their context regardless of source modality; this deliberate design isolates structured-output capability from raw vision or speech-processing quality, ensuring a fair, source-agnostic comparison. Our benchmark comprises \numtext{} text evaluation records derived from multi-hop QA (drawn from a 25,091-record full corpus), 209 image records from OCR-processed PDFs across seven document types (including multi-column layouts, dense tables, scanned historical documents, small-print text, and mathematical typesetting), and 115 audio records from the AMI corpus. Each record pairs a natural-language question with a JSON schema that the model must follow and a ground-truth answer verified against the source context. We evaluate 21 frontier and open-weight models across three source domains and seven metrics. Our results reveal a consistent pattern: models achieve near-perfect schema compliance, yet the best Value Accuracy (exact leaf-value match) reaches only 83.0\% on text, 67.2\% on images, and 23.7\% on audio, where longer context makes extraction substantially harder. We release the dataset, evaluation pipeline, and all related code.

\end{abstract}

\section{Introduction}
\label{sec:intro}
Most benchmarks for large language models (LLMs) evaluate reasoning, summarization, or code generation; the few that target structured output (JSONSchemaBench~\citep{geng2025jsonschemabench}, StructEval~\citep{li2025structeval}, and ExtractBench~\citep{extractbench2026}) stop at schema compliance or cover only a single source modality. In deployed systems, however, extracting structured records from unstructured and semi-structured content is among the most common uses of LLMs: parsing invoices, populating medical records, and converting documents to database entries. We refer to this capability as \emph{structured output}: given a context that may include OCR'd pages, meeting transcripts, or free-form text, the model must return JSON that both conforms to a target schema and carries values faithfully grounded in that context. Unlike open-ended generation, the task rewards deterministic extraction over creative completion.

The current evaluation landscape addresses parts of this problem in isolation. Some benchmarks test schema compliance~\citep{geng2025jsonschemabench} or format adherence~\citep{li2025structeval, zhou2023ifeval}. ExtractBench~\citep{extractbench2026} and LLMStructBench~\citep{tenckhoff2026llmstructbench} do evaluate value correctness, but are limited to a single source domain (images or text respectively) and do not test whether the same models maintain accuracy across different source types. A model that returns \texttt{"occupations":["American country music artist","composer"]} when the source text only supports \texttt{["country music artist","composer"]} passes every schema validator (see Table~\ref{tab:hallucinations}, group~a). The JSON is valid. The type is correct, but the data is wrong. A single error within a correctly formatted response can propagate silently through downstream systems. At the time of writing, we are not aware of any prior benchmark that jointly evaluates structured extraction across three source modalities (text, images, and audio), schema-specified JSON generation, and per-field exact value grounding at this scale under a unified evaluation framework.

This is not a hypothetical failure. In our evaluation of 21 state-of-the-art models, we find that models achieve high schema compliance on text, yet Value Accuracy (exact leaf-value match) is only 83.0\% for the best model. On images, it drops to 67.2\%. On audio, 23.7\%. The gap between ``produces valid JSON'' and ``produces correct JSON'' is real, large, and varies dramatically across both models and source domains.

We introduce \ours{} (The Structured Output Benchmark), which integrates multi-source extraction, value-level accuracy evaluation, and unified cross-source comparison into a single benchmark. Our contributions:
\begin{enumerate}[leftmargin=*, nosep]
    \item A \textbf{multi-source structured output benchmark} spanning three source modalities: native text, images, and audio conversations. We instantiate these modalities using HotpotQA~\citep{yang2018hotpotqa}, OCR-processed PDFs across seven document types, and meeting transcripts from the AMI corpus~\citep{ami_corpus}, all mapped into a unified evaluation schema.
    \item A \textbf{fine-grained evaluation methodology} comprising seven metrics (Table~\ref{tab:metric_defs}) that separately measure schema compliance, value accuracy, faithfulness, structural completeness, structural coverage, type safety, and strict precision, with aggregation.
    \item A \textbf{reproducible benchmark construction methodology} combining human authoring of schemas and ground truth with LLM cross-check (Gemini~2.5~Flash for per-record review) and LLM-as-a-judge~\citep{zheng2023llmasjudge} for multi-dimensional quality filtering, alongside Pydantic model validation and JSON Schema validation, designed for reproducible extension to new source domains.
    \item \textbf{Empirical findings} from evaluating 21 models across three source domains, revealing that model size does not predict structured output quality, Value Accuracy drops from 0.830 (text) to 0.672 (image) to 0.237 (audio), rankings shift across sources, and hallucinations inside JSON fields are harder to detect than in free-text responses because they look structurally correct.
\end{enumerate}

\begin{figure*}[t]
\centering
\includegraphics[width=\textwidth, trim=0 25 0 10]{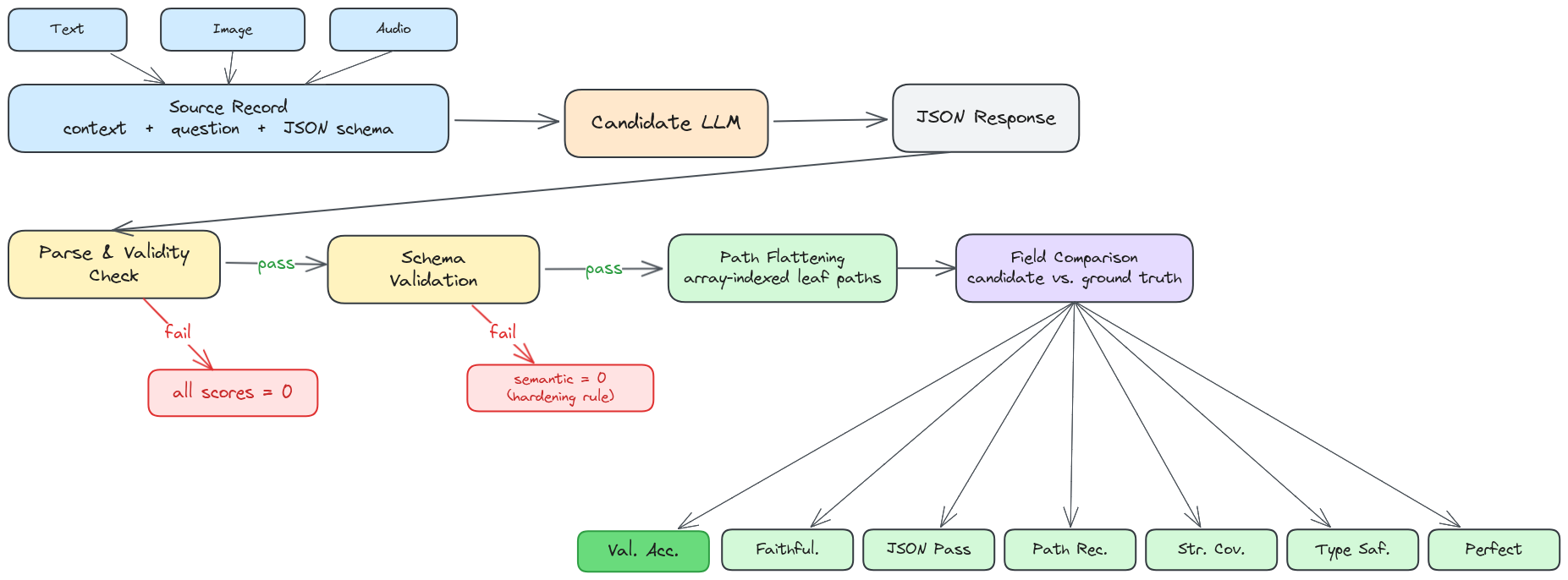}
\caption{\ours{} evaluation pipeline. Each source record (context, question, JSON schema) is submitted to the candidate model. The response is first checked for parse validity, then schema compliance: failures trigger the hardening rule (semantic scores zeroed). Passing responses are path-flattened to leaf nodes with concrete array indices, then compared field-by-field against ground truth to yield the seven evaluation metrics.}
\label{fig:eval_pipeline}
\end{figure*}

\section{Related Work}
\label{sec:related}
\subsection{Constrained Decoding}

A complementary line of work enforces grammar or schema compliance at generation time through \emph{constrained decoding}~\citep{willard2023efficient, dong2024xgrammar, zheng2024sglang}. However, these methods address only syntactic validity, not whether the extracted values are correct, and \citet{park2024grammar} show that naive grammar-constrained decoding distorts the LLM's distribution toward syntactically valid but semantically degraded outputs. \citet{tam2024letmespeak} further find that structured format constraints (JSON, XML, YAML) degrade LLM reasoning ability. These limitations motivate our benchmark's focus on value accuracy alongside schema compliance.

\subsection{Missing Dimension in Existing Benchmarks}

Several benchmarks evaluate aspects of structured generation. Table~\ref{tab:benchmark_comparison} (Appendix~\ref{app:comparison}) provides a systematic comparison across key axes: source domain coverage, value accuracy, multi-hop reasoning, and per-field grounding. \textbf{JSONSchemaBench}~\citep{geng2025jsonschemabench} evaluates ${\sim}10{,}000$ real-world JSON schemas across six constrained decoding frameworks, focusing on schema coverage and generation efficiency rather than content accuracy. \textbf{StructEval}~\citep{li2025structeval} tests 18 structured formats across 44 task types, finding that even o1-mini reaches only a 75.58 average score. \textbf{DeepJSONEval}~\citep{deepjsoneval2025} measures performance on deeply nested JSON structures and reports significant degradation as nesting depth grows, but tests only structural depth, not value accuracy. \textbf{LLMStructBench}~\citep{tenckhoff2026llmstructbench} benchmarks 22 LLMs across parsing scenarios with five prompting strategies, finding that prompting strategy matters more than model size and that semantic errors persist even when structural validity is achieved.

Most closely related, \textbf{ExtractBench}~\citep{extractbench2026} pairs 35 PDF documents with JSON schemas and gold labels spanning 12,867 fields, finding frontier models achieve only 4.6\% field-level pass rate. While ExtractBench tests document-to-JSON extraction, it is limited to a single source domain and does not require multi-hop reasoning.

On the evaluation methodology side, \textbf{STED}~\citep{wang2025sted} introduces Semantic Tree Edit Distance for comparing JSON outputs with a consistency scoring framework. \citet{chen2023unified_metrics} propose a unified framework for evaluation metrics across structured prediction tasks. In the instruction-following space, \textbf{IFEval}~\citep{zhou2023ifeval}, \textbf{FollowBench}~\citep{jiang2024followbench}, and \textbf{InFoBench}~\citep{qin2024infobench} evaluate various forms of constraint adherence, while \citet{pyatkin2025ifbench} show that models overfit existing benchmarks. Our work extends the verifiable evaluation principle to per-field JSON compliance and value accuracy across three source domains.

\subsection{Structured Output Gaps in General Multi-Modal Benchmarks}

General multi-modal benchmarks (MMBench, MMMU)~\citep{liu2024mmbench, yue2024mmmu} target broad reasoning, while document-understanding benchmarks (DocVQA, ChartQA, InfographicVQA, OCRBench~v2)~\citep{mathew2021docvqa, masry2022chartqa, mathew2022infographicvqa, fu2025ocrbenchv2} and audio benchmarks (AudioBench)~\citep{wang2024audiobench} address modality-specific QA. None of these requires schema-compliant JSON output with grounded per-field value accuracy across modalities. \ours{} closes this gap, enabling direct comparison of model capabilities across native text, images, and audio after text normalization.

\subsection{Faithfulness, Tool Use, and Dataset Construction}

Our per-field evaluation extends FActScore's atomic-verification approach~\citep{min2023factscore} from free text to structured outputs, with the source context taking the role of the knowledge base; we use this lens to characterise the value-level hallucinations catalogued in the broader hallucination taxonomy of \citet{huang2024hallucination}. Tool-use benchmarks~\citep{patil2024gorilla, yan2024bfcl, qin2024toolllm} test API call formatting against documentation; we instead test extraction from semi-structured context (OCR transcripts, meeting logs, multi-hop Wikipedia passages). Our text source is HotpotQA~\citep{yang2018hotpotqa}; MuSiQue~\citep{trivedi2022musique} and 2WikiMultihopQA~\citep{ho2020wikimultihopqa} are complementary multi-hop resources we leave for future expansion. Our construction pipeline uses an LLM judge~\citep{zheng2023llmasjudge} for quality filtering, following the reproducible-pipeline conventions of DataDreamer~\citep{patel2024datadreamer}.

\section{Methodology}
\label{sec:construction}
We build \ours{} by processing three open-source datasets corresponding to our three source modalities: HotpotQA~\citep{yang2018hotpotqa} for native text, olmOCR-bench~\citep{poznanski2025olmocr} for images, and the AMI Meeting Corpus~\citep{ami_corpus} for audio conversations. For each record, the JSON schema and ground-truth structured output are produced through human review of the source content, then cross-checked and verified by an LLM reviewer (Gemini~2.5~Flash) that flags missing fields, type mismatches, and ungrounded values. The image source (209 records) and audio source (115 records) were reviewed record-by-record; for the text source, five independent 100-record random samples estimated a $\sim$3\% residual error rate after the human-and-LLM pass, with all identified errors manually corrected. Full per-source construction details and the authoring-and-verification pipeline are in Appendix~\ref{app:construction}. Every record follows:
\[
\texttt{record} = \bigl\{\texttt{input}: \{c, q, s\},\; \texttt{output}: \{g, v\}\bigr\}
\]
where $c$ is context, $q$ is a question, $s$ is a JSON schema, $g$ is the ground-truth structured output, and $v$ is the validated output after schema alignment. This shared schema enables a single evaluation script to score all source domains uniformly.

\textbf{Input representation.} A key design choice in \ours{} is that all models receive their context as text, regardless of input source. Image records are supplied as OCR-rendered markdown transcripts; audio records are supplied as timestamped, speaker-labelled conversation transcripts. For example, an image record may consist of OCR text from an invoice or report with headers, tables, and totals, while an audio record may consist of a meeting transcript with timestamps and speaker turns. This is intentional: our goal is to benchmark \emph{structured-output capability}, the ability to extract grounded, schema-compliant values from an accurate textual representation, not raw vision or ASR processing capability. Passing raw images or audio would introduce a confound: models with stronger native vision or speech stacks would outperform on structured extraction for the wrong reason. By separating the OCR/transcription step (performed once, validated for quality) from the structured generation step, we ensure rankings reflect the capability \ours{} actually measures. End-to-end evaluation with native image and audio inputs (bypassing the OCR/transcription step) is planned as a future extension (\S\ref{sec:limitations}).

Each record's schema is categorized by complexity: \textbf{medium} (nested objects or arrays of scalars, depth~2) or \textbf{hard} (arrays of objects, or depth~$\geq$3). Across all source domains, the majority of schemas are hard (61\% text, 88\% image, 98\% audio; Appendix~\ref{app:dataset_stats}); representative hard examples are shown in Appendix~\ref{app:example}.

As a concrete example, one text record contains two Wikipedia paragraphs: one mentions Hal Ashby was an \emph{American} film director, another states Ciro Ippolito is \emph{Italian}. The question asks: \emph{``Are Hal Ashby and Ciro Ippolito of the same nationality?''} The model must return JSON conforming to this schema:

\begin{lstlisting}[basicstyle=\ttfamily\scriptsize, frame=single, columns=fullflexible]
{"type": "object",
 "required": ["directors", "are_same_nationality", "reasoning"],
 "properties": {
   "directors": {"type": "array", "items": {"type": "object",
     "properties": {"name": {"type": "string"}, "nationality": {"type": "string"}},
     "required": ["name", "nationality"]}},
   "are_same_nationality": {"type": "boolean"},
   "reasoning": {"type": "string"}}}
\end{lstlisting}
The ground-truth answer:
\begin{lstlisting}[basicstyle=\ttfamily\scriptsize, frame=single, columns=fullflexible]
{"directors": [{"name": "Hal Ashby", "nationality": "American"},
               {"name": "Ciro Ippolito", "nationality": "Italian"}],
 "are_same_nationality": false,
 "reasoning": "Hal Ashby was American, Ciro Ippolito is Italian."}
\end{lstlisting}
This record requires multi-hop reasoning (two paragraphs), nested structure (array of objects), and mixed types (string, boolean). A model returning \texttt{"United States"} instead of \texttt{"American"} would score 0 on that leaf in Value Accuracy despite being semantically plausible.

\section{Evaluation}
\label{sec:eval}
\subsection{Base Metrics}

Figure~\ref{fig:eval_pipeline} shows the end-to-end evaluation pipeline. Each record is scored by comparing the candidate response against ground truth through JSON path flattening. Let $\mathcal{G}$ and $\mathcal{P}$ denote the flattened leaf-path maps of ground truth and prediction, with $G = \text{paths}(\mathcal{G})$, $P = \text{paths}(\mathcal{P})$, and $O = G \cap P$. We report seven metrics (Table~\ref{tab:metric_defs}), all defined below with key equations. Full mathematical definitions for all seven are in Appendix~\ref{app:metrics}.

\begin{table}[h]
\centering
\caption{Summary of the seven evaluation metrics. All are per-record, then averaged across the set.}
\label{tab:metric_defs}
\small
\begin{tabular}{@{}llp{6.2cm}@{}}
\toprule
\textbf{Metric} & \textbf{Range} & \textbf{What it measures} \\
\midrule
JSON Pass Rate & $\{0,1\}$ & Parse + structured root + schema validates \\
Faithfulness & $[0,1]$ & Soft value match (token F1) \\
Path Recall & $[0,1]$ & Structural completeness \\
Structure Coverage & $[0,1]$ & Structural precision$\times$recall (F1) \\
Type Safety & $[0,1]$ & JSON type correctness \\
Perfect Response & $\{0,1\}$ & Exact full-object match \\
\textbf{Value Accuracy} & $[0,1]$ & \textbf{Exact leaf-value match (primary)} \\
\bottomrule
\end{tabular}
\end{table}

\textbf{Value Accuracy} is the primary metric. It measures the fraction of ground-truth leaf paths where the predicted value exactly matches:
\begin{equation}
\text{Value Accuracy} = \frac{|\{p \in G : \mathcal{G}[p] = \mathcal{P}[p]\}|}{|G|}
\end{equation}
Because paths include concrete array indices (e.g., \texttt{items.0.name}, \texttt{items.1.name}), Value Accuracy is inherently sensitive to array element ordering: a model that returns the correct values in the wrong order scores 0 on those leaves. This is intentional: for ordered sequences such as receipt line items, speaker turns, or timeline events, positional correctness is part of the extraction task.

\textbf{Faithfulness Score} complements Value Accuracy with a softer, partial-credit measure via token-level F1. Where Value Accuracy is binary per leaf (exact match or zero), Faithfulness gives partial credit for overlapping tokens (e.g., ``American film director'' vs ``American'' would score high on Faithfulness but 0 on Value Accuracy). For each ground-truth leaf $p \in G$, we compute F1 between the tokenized, normalized (lowercased, articles removed, punctuation stripped) strings of $\mathcal{G}[p]$ and $\mathcal{P}[p]$. Missing paths contribute $\text{F1} = 0$. The metric is the mean over all leaves:
\begin{equation}
\text{Faithfulness} = \frac{1}{|G|}\sum_{p \in G} \text{F1}\bigl(\text{toks}(\mathcal{G}[p]),\; \text{toks}(\mathcal{P}[p])\bigr)
\end{equation}

\textbf{Structure Coverage} is the F1 over path sets, capturing how well the predicted structure aligns with the expected structure:
\begin{equation}
\text{Structure Coverage} = \frac{2 \cdot \frac{|O|}{|P|} \cdot \frac{|O|}{|G|}}{\frac{|O|}{|P|} + \frac{|O|}{|G|}}
\end{equation}

\textbf{JSON Pass Rate} is a strict binary gate: 1 if the response parses as valid JSON, has a structured root (\texttt{dict} or \texttt{list}), and passes \texttt{jsonschema.validate} against the target schema. \textbf{Path Recall} is $|O|/|G|$, the fraction of ground-truth paths present in prediction. \textbf{Type Safety} is the fraction of predicted leaf values whose JSON type matches the schema-expected type (with array indices wildcarded). \textbf{Perfect Response} is 1 iff the canonicalized (recursively key-sorted) prediction equals the canonicalized ground truth exactly.

\subsection{Hardening and Coverage Gates}

Raw semantic metrics (Value Accuracy, Faithfulness, Path Recall, Structure Coverage) are multiplied by a hardening factor $h$ that gates on structural correctness: if a response fails to parse, lacks a structured root, or violates the schema, all semantic scores are driven to zero. For text, we additionally apply a hard coverage gate (floor $= 0.95$); for image and audio sources, a softer gate (floor $= 0.90$) that gives partial credit under noisy OCR or long transcripts. Full equations are in Appendix~\ref{app:metrics}.

\subsection{Evaluation Categories}

We group the seven reported metrics plus one parse-only auxiliary sub-check (JSON Parse Success) into five categories that distinguish long-context extraction, schema handling, multi-context linking, output reliability, and exact-match strictness: \emph{Long Context Extraction} (Value Accuracy, Faithfulness, Path Recall), \emph{Complex Schema Handling} (JSON Pass Rate, Structure Coverage, Type Safety), \emph{Multi-Context Linking} (Value Accuracy, Faithfulness), \emph{Output Contract Reliability} (JSON Parse Success, JSON Pass Rate, Type Safety), and \emph{Strict Precision} (Perfect Response Rate). JSON Parse Success is the parse-only sub-check; including it alongside JSON Pass Rate gives partial credit when output parses but fails schema validation. Full definitions are in Appendix~\ref{app:categories}.

\subsection{Aggregation}

Category scores per record are arithmetic means of the metrics assigned to that category (e.g., Value Accuracy, Faithfulness, and Path Recall for Long Context Extraction). Unless otherwise stated, the main tables report schema-complexity-weighted means (easy${}=1.0$, medium${}=2.0$, hard${}=3.0$), with the weight determined by each record's schema complexity. The weights reflect that production extraction tasks overwhelmingly involve complex, nested schemas: a model that handles only flat objects but fails on arrays of objects is less useful than one that handles both, and the weighted score should reflect this.
\section{Experimental Setup}
\label{sec:setup}
\subsection{Models Evaluated}

We evaluate 21 models (20 on audio; Phi-4 excluded due to its 16K context limit) spanning frontier API providers and open-weight models served via vLLM~\citep{kwon2023vllm}, ranging from 8B to 358B parameters.

\subsection{Inference Configuration}

All models are evaluated at temperature 0.0 (greedy decoding) with maximum output length of 2,048 tokens. Open-weight models use vLLM~\citep{kwon2023vllm} with tensor parallelism. Each model receives context, question, and JSON schema as input and must return a conforming JSON response.

Reasoning is disabled where the provider supports it. Full disable is not possible for three models: GPT-5, GPT-5-Mini (minimum reasoning effort), and DS-R1-Distill-32B (reasoning-distilled; chain-of-thought intrinsic to training). We run them in their lowest-reasoning configuration. Structured extraction in our setting is closer to retrieval-plus-transpilation than to open-ended problem solving: the relevant ground truth is already present in the provided context. Evaluating in non-reasoning mode isolates extraction capability from compute budget and reflects how these models are typically deployed for backend extraction pipelines.

\section{Results and Discussion}
\label{sec:results}
\subsection{Unified Overall Leaderboard}

Before per-modality analysis, we report a single unified leaderboard that includes all 21 models and aggregates base metrics across text, image, and audio. For each metric $k$, we compute a schema-complexity-weighted aggregate over the available settings for that model:
\begin{equation}
\bar{m}_k=\frac{\sum_{u \in \mathcal{U}_k} W_u m^{(w)}_{k,u}}{\sum_{u \in \mathcal{U}_k} W_u},
\end{equation}
where $\mathcal{U}_k$ is the set of settings with valid scores for metric $k$, $m^{(w)}_{k,u}$ is the within-source metric, and $W_u$ is the total schema-complexity weight for that source. In our evaluation, $(W_t,W_i,W_a)=(13054,602,343)$ for text, image, and audio respectively. For Perfect Response (not reported for audio), we aggregate over text+image only:
\begin{equation}
\bar{m}_{\text{perfect}}=\frac{W_t m^{(w)}_{\text{perfect},t}+W_i m^{(w)}_{\text{perfect},i}}{W_t+W_i}.
\end{equation}
The raw overall score is the arithmetic mean of aggregated metrics:
\begin{equation}
\text{Overall (Raw)}=\frac{1}{|K|}\sum_{k \in K} \bar{m}_k,
\end{equation}
with $K=\{\text{truth, faithfulness, json pass, path recall, structure coverage, type safety, perfect}\}$ and $|K|$ equal to the number of available metrics for that model. To make rankings comparable under partial runs, we apply a coverage factor:
\begin{equation}
\text{Coverage}=\frac{n_{\text{eval}}}{5000+209+115}, \quad
\text{Overall (Adj.)}=\text{Overall (Raw)}\times \text{Coverage}.
\end{equation}
Table~\ref{tab:overall_leaderboard} is sorted by Overall (Adj.). Across models, JSON Pass remains uniformly high in the overall aggregation, while Value Accuracy trails it by a visibly larger margin. This reinforces the central distinction measured by \ours{}: producing schema-valid JSON is substantially easier than extracting the correct grounded values inside that JSON.

\begin{table}[t]
\centering
\caption{Unified overall leaderboard for all 21 models, sorted by coverage-adjusted Overall score.}
\label{tab:overall_leaderboard}
\scriptsize
\setlength{\tabcolsep}{2.5pt}
\begin{tabular}{@{}lrrrrrrrr@{}}
\toprule
\textbf{Model} & \rotatebox{65}{\scriptsize \textbf{Overall}} & \rotatebox{65}{\scriptsize Val.~Acc.} & \rotatebox{65}{\scriptsize Faithful.} & \rotatebox{65}{\scriptsize JSON Pass} & \rotatebox{65}{\scriptsize Path Rec.} & \rotatebox{65}{\scriptsize Str.~Cov.} & \rotatebox{65}{\scriptsize Type Saf.} & \rotatebox{65}{\scriptsize Perfect} \\
\midrule
GPT-5.4                & \textbf{0.870} & 0.798 & \textbf{0.869} & \textbf{0.993} & \textbf{0.988} & \textbf{0.981} & \textbf{0.993} & 0.469 \\
GLM-4.7                & 0.861 & \textbf{0.804} & 0.868 & 0.965 & 0.959 & 0.957 & 0.965 & \textbf{0.508} \\
Qwen3.5-35B            & 0.861 & 0.801 & 0.863 & 0.969 & 0.962 & 0.960 & 0.969 & 0.500 \\
Gemini-2.5-Flash       & 0.860 & 0.796 & 0.856 & 0.972 & 0.967 & 0.961 & 0.972 & 0.498 \\
Qwen3-235B             & 0.857 & 0.786 & 0.854 & 0.978 & 0.970 & 0.968 & 0.978 & 0.463 \\
Interfaze-Beta         & 0.855 & 0.795 & 0.858 & 0.967 & 0.962 & 0.957 & 0.967 & 0.480 \\
Claude-Sonnet-4.6      & 0.854 & 0.779 & 0.858 & 0.979 & 0.975 & 0.969 & 0.979 & 0.442 \\
GPT-4.1                & 0.850 & 0.783 & 0.853 & 0.969 & 0.963 & 0.959 & 0.969 & 0.454 \\
GPT-5                  & 0.849 & 0.769 & 0.859 & 0.983 & 0.978 & 0.972 & 0.983 & 0.398 \\
Gemma-3-27B            & 0.847 & 0.777 & 0.842 & 0.969 & 0.961 & 0.958 & 0.969 & 0.454 \\
Qwen3-30B              & 0.842 & 0.753 & 0.832 & 0.983 & 0.974 & 0.970 & 0.983 & 0.401 \\
Nemotron-3-Nano-30B    & 0.841 & 0.747 & 0.817 & 0.987 & 0.975 & 0.971 & 0.987 & 0.400 \\
GPT-5-Mini             & 0.835 & 0.751 & 0.837 & 0.972 & 0.966 & 0.960 & 0.972 & 0.388 \\
Gemma-4-31B            & 0.833 & 0.778 & 0.843 & 0.943 & 0.934 & 0.934 & 0.943 & 0.461 \\
Gemini-3-Flash-Preview & 0.833 & 0.773 & 0.831 & 0.939 & 0.935 & 0.929 & 0.939 & 0.484 \\
Schematron-8B          & 0.832 & 0.731 & 0.807 & 0.987 & 0.976 & 0.969 & 0.987 & 0.370 \\
IBM-Granite-4.0        & 0.832 & 0.736 & 0.812 & 0.983 & 0.965 & 0.967 & 0.983 & 0.381 \\
Phi-4                  & 0.831 & 0.787 & 0.849 & 0.969 & 0.961 & 0.961 & 0.969 & 0.452 \\
DS-R1-Distill-32B      & 0.827 & 0.747 & 0.819 & 0.960 & 0.945 & 0.947 & 0.960 & 0.411 \\
Ministral-3-14B        & 0.778 & 0.700 & 0.773 & 0.906 & 0.898 & 0.896 & 0.906 & 0.368 \\
GPT-OSS-20B            & 0.732 & 0.667 & 0.730 & 0.845 & 0.838 & 0.836 & 0.845 & 0.362 \\
\bottomrule
\end{tabular}
\end{table}

\subsection{JSON Pass vs Val Acc}
\begin{figure}[t]
\centering
\includegraphics[width=0.85\textwidth]{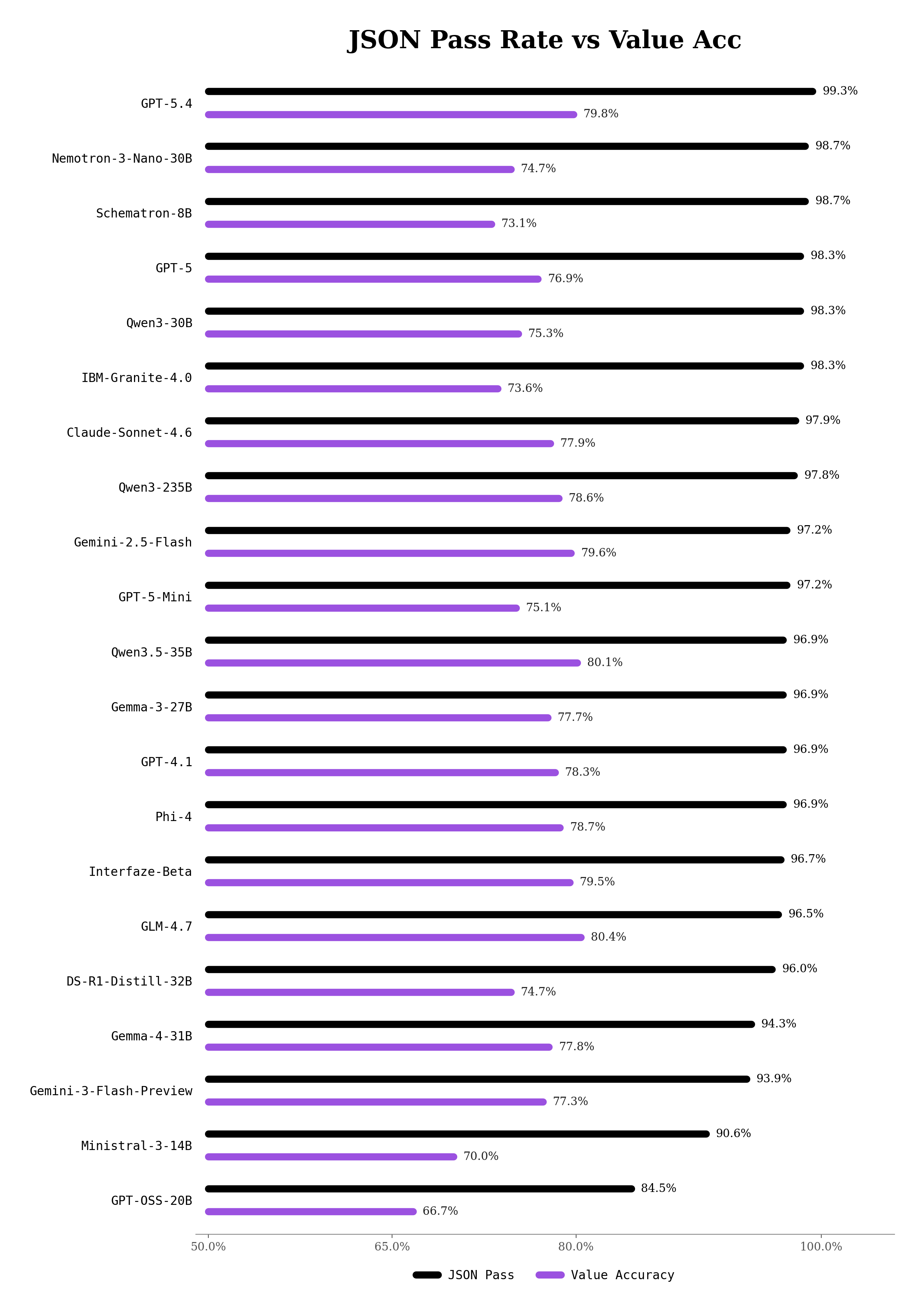}                                  
\caption{JSON Pass Rate (black) vs Value Accuracy (purple) in percentages across all 21 evaluated models, sorted by JSON Pass Rate. Every model exceeds 84\% JSON Pass, yet no model surpasses 80.4\% Value Accuracy. The gap between the two bars is the structured-output hallucination that schema-only benchmarks do not measure.}                               \label{fig:json_vs_val}
\end{figure}

As shown in Figure~\ref{fig:json_vs_val}, we observe a significant drop across all 21 models between JSON Pass Rate (a metric that all existing structured-output benchmarks measure) and Value Accuracy (whether the values produced are actually correct). Across the leaderboard, the gap is consistently 15--25 percentage points: every model produces
nearly perfect JSON, yet a sizeable fraction of the leaf values inside that JSON are wrong. This is a key hallucination class that gets lost in many data pipelines which rely on LLMs to produce accurate structured data from raw information; because the response parses, validates, and looks correct, the error propagates silently into downstream systems.
                                     
More importantly, if we read Value Accuracy independently, the leaderboard ranking changes significantly. Open-source mid-sized models such as Qwen3.5-35B (0.801) outperform all closed-source SoTA models in our evaluation, including GPT-5.4 (0.798), Gemini-2.5-Flash (0.796), and Claude-Sonnet-4.6 (0.779). We also note that we do not see significant improvement in Value Accuracy across the GPT model family from its predecessor: GPT-5.4 exceeds GPT-4.1 by only $\sim$0.02 in Value Accuracy (0.798 vs 0.783) and GPT-5 by 0.029 (0.798 vs 0.769), despite the substantial gap in headline reasoning benchmarks between these generations. This suggests that scaling compute alone has not closed the value-accuracy gap, and that structured-output capability is largely orthogonal to the abilities measured by traditional reasoning and chat benchmarks.

\subsection{Structured Decoding Ablation}
\label{sec:structured_decoding_ablation}

Table~\ref{tab:structured_decoding_ablation} shows the difference between the main setting and explicit schema-constrained decoding. The baseline is the main evaluation setting: greedy decoding with the schema provided in the prompt. The constrained setting keeps the same prompt and decoding setup, but additionally passes the schema to the provider for enforcement. Value Accuracy changes only slightly, from $-0.007$ to $+0.033$, so this choice does not affect the main accuracy conclusions.

\begin{table}[h]
\centering
\caption{Ablation isolating explicit schema-constrained decoding (audio source, $n=115$). Three models tested.}
\label{tab:structured_decoding_ablation}
\scriptsize
\setlength{\tabcolsep}{4pt}
\begin{tabular}{@{}lrrrr@{}}
\toprule
\textbf{Model} & \textbf{Val.~Acc. Base} & \textbf{Val.~Acc. +Schema} & \textbf{JSON Pass Base} & \textbf{JSON Pass +Schema} \\
\midrule
Gemini-2.5-Flash & 0.237 & 0.270 & 0.860 & 0.956 \\
Gemini-3-Flash-Preview & 0.190 & 0.212 & 0.773 & 0.869 \\
GPT-5.4          & 0.180 & 0.173 & 0.869 & 0.808 \\
\bottomrule
\end{tabular}
\end{table}
\FloatBarrier

\subsection{Text Results}

Table~\ref{tab:results} presents schema-complexity-weighted per-metric scores for all 21 valid models on text (\numtest{} records), sorted by Value Accuracy (rightmost column, our primary metric). GLM-4.7 leads at 0.830, followed by Qwen3.5-35B (0.828) and GPT-5.4 (0.825). The structural metrics (JSON Pass, Path Recall, Type Safety) cluster above .95 for most models, yet Value Accuracy spans a 13.7-point range (0.693 to 0.830), showing that value accuracy varies far more than structural compliance.

\begin{table}[h]
\centering
\caption{Text results ($n = 5{,}000$, 21 valid models). Best per metric in \textbf{bold}.}
\label{tab:results}
\scriptsize
\setlength{\tabcolsep}{2.5pt}
\begin{tabular}{@{}lrrrrrr|r@{}}
\toprule
\textbf{Model} & \rotatebox{65}{\scriptsize Faithful.} & \rotatebox{65}{\scriptsize JSON Pass} & \rotatebox{65}{\scriptsize Path Rec.} & \rotatebox{65}{\scriptsize Str.~Cov.} & \rotatebox{65}{\scriptsize Type Saf.} & \rotatebox{65}{\scriptsize Perfect} & \rotatebox{65}{\scriptsize \textbf{Val.~Acc.}} \\
\midrule
GLM-4.7                & 0.884 & 0.972 & 0.967 & 0.967 & 0.972 & \textbf{0.526} & \textbf{0.830} \\
Qwen3.5-35B            & 0.881 & 0.974 & 0.970 & 0.968 & 0.974 & 0.519 & 0.828 \\
GPT-5.4                & \textbf{0.887} & \textbf{0.999} & \textbf{0.996} & \textbf{0.992} & \textbf{0.999} & 0.486 & 0.825 \\
Gemini-2.5-Flash       & 0.874 & 0.983 & 0.980 & 0.975 & 0.983 & 0.515 & 0.822 \\
Interfaze-Beta         & 0.876 & 0.975 & 0.973 & 0.968 & 0.975 & 0.497 & 0.821 \\
Qwen3-235B             & 0.870 & 0.982 & 0.977 & 0.974 & 0.982 & 0.481 & 0.811 \\
GPT-4.1                & 0.872 & 0.974 & 0.971 & 0.968 & 0.974 & 0.470 & 0.811 \\
Claude-Sonnet-4.6      & 0.876 & 0.984 & 0.981 & 0.977 & 0.984 & 0.459 & 0.809 \\
Gemma-3-27B            & 0.861 & 0.975 & 0.971 & 0.967 & 0.975 & 0.471 & 0.803 \\
Gemini-3-Flash-Preview & 0.851 & 0.951 & 0.949 & 0.944 & 0.951 & 0.501 & 0.800 \\
Gemma-4-31B            & 0.855 & 0.945 & 0.939 & 0.939 & 0.945 & 0.477 & 0.798 \\
Phi-4                  & 0.855 & 0.974 & 0.966 & 0.966 & 0.974 & 0.468 & 0.798 \\
GPT-5                  & 0.878 & 0.988 & 0.983 & 0.980 & 0.988 & 0.412 & 0.795 \\
GPT-5-Mini             & 0.857 & 0.977 & 0.973 & 0.969 & 0.977 & 0.401 & 0.779 \\
Qwen3-30B              & 0.849 & 0.988 & 0.983 & 0.978 & 0.988 & 0.414 & 0.778 \\
Nemotron-3-Nano-30B    & 0.835 & 0.991 & 0.982 & 0.980 & 0.991 & 0.416 & 0.774 \\
DS-R1-Distill-32B      & 0.838 & 0.964 & 0.955 & 0.955 & 0.964 & 0.427 & 0.773 \\
IBM-Granite-4.0        & 0.829 & 0.985 & 0.974 & 0.974 & 0.985 & 0.397 & 0.761 \\
Schematron-8B          & 0.824 & 0.990 & 0.982 & 0.977 & 0.990 & 0.384 & 0.754 \\
Ministral-3-14B        & 0.788 & 0.909 & 0.904 & 0.902 & 0.909 & 0.382 & 0.724 \\
GPT-OSS-20B            & 0.752 & 0.858 & 0.853 & 0.851 & 0.858 & 0.376 & 0.693 \\
\bottomrule
\end{tabular}
\end{table}

\subsection{Image and Audio Results}

Tables~\ref{tab:image_results} and~\ref{tab:audio_results} present results for images (209 records, 21 valid models) and audio (115 records, 20 models). Value Accuracy scores drop sharply across modalities: the best image model (Gemma-4-31B) reaches 0.672, roughly 81\% of the text leader. Audio is harder still: Gemini~2.5~Flash leads at 0.237, less than one-third of the text ceiling. JSON Pass Rates remain high across all three settings (most models $\geq$80\% even on audio), confirming that structural compliance is not the bottleneck. The challenge is accurate value extraction from noisy OCR and complex multi-speaker transcripts.

\begin{table}[t]
\centering
\caption{Image results ($n\!=\!209$, 21 valid models), sorted by Value Accuracy. Best per metric in \textbf{bold}.}
\label{tab:image_results}
\scriptsize
\setlength{\tabcolsep}{2.5pt}
\begin{tabular}{@{}lrrrrrr|r@{}}
\toprule
\textbf{Model} & \rotatebox{65}{\scriptsize Faithful.} & \rotatebox{65}{\scriptsize JSON Pass} & \rotatebox{65}{\scriptsize Path Rec.} & \rotatebox{65}{\scriptsize Str.~Cov.} & \rotatebox{65}{\scriptsize Type Saf.} & \rotatebox{65}{\scriptsize Perfect} & \rotatebox{65}{\scriptsize \textbf{Val.~Acc.}} \\
\midrule
Gemma-4-31B            & \textbf{0.837} & \textbf{0.995} & \textbf{0.963} & \textbf{0.957} & \textbf{0.995} & 0.113 & \textbf{0.672} \\
GLM-4.7                & 0.747 & 0.849 & 0.837 & 0.830 & 0.849 & 0.113 & 0.575 \\
Gemini-2.5-Flash       & 0.691 & 0.786 & 0.776 & 0.766 & 0.786 & \textbf{0.130} & 0.559 \\
Interfaze-Beta         & 0.721 & 0.852 & 0.837 & 0.827 & 0.852 & 0.115 & 0.559 \\
Phi-4                  & 0.725 & 0.875 & 0.843 & 0.845 & 0.875 & 0.100 & 0.557 \\
Qwen3-235B             & 0.742 & 0.907 & 0.885 & 0.881 & 0.907 & 0.078 & 0.552 \\
Gemma-3-27B            & 0.720 & 0.880 & 0.857 & 0.854 & 0.880 & 0.088 & 0.552 \\
Qwen3.5-35B            & 0.717 & 0.854 & 0.834 & 0.829 & 0.854 & 0.098 & 0.550 \\
GPT-5.4                & 0.745 & 0.920 & 0.912 & 0.878 & 0.920 & 0.101 & 0.546 \\
Schematron-8B          & 0.729 & 0.917 & 0.888 & 0.878 & 0.917 & 0.078 & 0.541 \\
GPT-4.1                & 0.714 & 0.867 & 0.852 & 0.836 & 0.867 & 0.103 & 0.526 \\
GPT-5                  & 0.737 & 0.932 & 0.917 & 0.893 & 0.932 & 0.090 & 0.526 \\
Gemini-3-Flash-Preview & 0.653 & 0.764 & 0.753 & 0.740 & 0.764 & 0.125 & 0.525 \\
Qwen3-30B              & 0.710 & 0.884 & 0.854 & 0.849 & 0.884 & 0.103 & 0.521 \\
IBM-Granite-4.0        & 0.711 & 0.942 & 0.886 & 0.897 & 0.942 & 0.045 & 0.517 \\
DS-R1-Distill-32B      & 0.687 & 0.879 & 0.835 & 0.844 & 0.879 & 0.065 & 0.514 \\
Nemotron-3-Nano-30B    & 0.691 & 0.904 & 0.874 & 0.862 & 0.904 & 0.053 & 0.501 \\
GPT-5-Mini             & 0.699 & 0.910 & 0.888 & 0.865 & 0.910 & 0.086 & 0.484 \\
Claude-Sonnet-4.6      & 0.714 & 0.887 & 0.877 & 0.855 & 0.887 & 0.071 & 0.466 \\
Ministral-3-14B        & 0.652 & 0.824 & 0.795 & 0.797 & 0.824 & 0.061 & 0.454 \\
GPT-OSS-20B            & 0.565 & 0.746 & 0.721 & 0.709 & 0.746 & 0.073 & 0.427 \\
\bottomrule
\end{tabular}
\end{table}

\begin{table}[t]
\centering
\caption{Audio results ($n\!=\!115$, all 20 models), sorted by Value Accuracy. Perfect Response Rate is $\leq$0.9\% for all models and omitted.}
\label{tab:audio_results}
\scriptsize
\setlength{\tabcolsep}{2.5pt}
\begin{tabular}{@{}lrrrrr|r@{}}
\toprule
\textbf{Model} & \rotatebox{65}{\scriptsize Faithful.} & \rotatebox{65}{\scriptsize JSON Pass} & \rotatebox{65}{\scriptsize Path Rec.} & \rotatebox{65}{\scriptsize Str.~Cov.} & \rotatebox{65}{\scriptsize Type Saf.} & \rotatebox{65}{\scriptsize \textbf{Val.~Acc.}} \\
\midrule
Gemini-2.5-Flash       & \textbf{0.464} & 0.860 & 0.816 & 0.791 & 0.860 & \textbf{0.237} \\
GLM-4.7                & 0.454 & 0.904 & 0.844 & 0.830 & 0.904 & 0.219 \\
Qwen3-235B             & 0.433 & 0.983 & 0.866 & \textbf{0.890} & 0.983 & 0.217 \\
Qwen3.5-35B            & 0.440 & 0.965 & 0.868 & 0.872 & 0.965 & 0.215 \\
Interfaze-Beta         & 0.409 & 0.843 & 0.751 & 0.759 & 0.843 & 0.205 \\
Claude-Sonnet-4.6      & 0.420 & 0.956 & \textbf{0.923} & 0.861 & 0.956 & 0.201 \\
Qwen3-30B              & 0.394 & \textbf{0.991} & 0.878 & 0.886 & \textbf{0.991} & 0.197 \\
GPT-4.1                & 0.380 & 0.930 & 0.876 & 0.836 & 0.930 & 0.191 \\
Ministral-3-14B        & 0.386 & 0.904 & 0.822 & 0.820 & 0.904 & 0.191 \\
Gemini-3-Flash-Preview & 0.367 & 0.773 & 0.697 & 0.700 & 0.773 & 0.190 \\
GPT-5                  & 0.364 & 0.913 & 0.870 & 0.803 & 0.913 & 0.188 \\
DS-R1-Distill-32B      & 0.333 & 0.965 & 0.759 & 0.818 & 0.965 & 0.180 \\
GPT-5.4                & 0.370 & 0.869 & 0.849 & 0.757 & 0.869 & 0.180 \\
Gemma-4-31B            & 0.375 & 0.773 & 0.678 & 0.700 & 0.773 & 0.178 \\
Gemma-3-27B            & 0.349 & 0.886 & 0.748 & 0.785 & 0.886 & 0.173 \\
IBM-Granite-4.0        & 0.324 & 0.956 & 0.779 & 0.825 & 0.956 & 0.161 \\
Schematron-8B          & 0.323 & 0.965 & 0.892 & 0.827 & 0.965 & 0.156 \\
Nemotron-3-Nano-30B    & 0.319 & \textbf{0.991} & 0.874 & 0.852 & \textbf{0.991} & 0.152 \\
GPT-5-Mini             & 0.330 & 0.886 & 0.834 & 0.788 & 0.886 & 0.150 \\
GPT-OSS-20B            & 0.185 & 0.539 & 0.499 & 0.468 & 0.539 & 0.088 \\
\bottomrule
\end{tabular}
\end{table}

\subsection{Key Findings}

Eight patterns emerge from the results:

\begin{enumerate}[leftmargin=*, nosep, label=\arabic*.]
\item Valid JSON does not mean correct JSON. GPT-5.4 reaches 99.97\% JSON Pass Rate, yet its Perfect Response Rate is only 0.486, about a 51-point gap. GLM-4.7, the Value Accuracy leader, reaches 0.526 perfect responses despite 0.972 JSON Pass Rate. This gap is precisely what prior benchmarks like JSONSchemaBench~\citep{geng2025jsonschemabench} do not measure.
\item Schema constraints have a smaller effect on grounded extraction than on structural reliability, on average. Table~\ref{tab:structured_decoding_ablation} shows the effect varies by model: Gemini-2.5-Flash and Gemini-3-Flash-Preview see JSON Pass rise; GPT-5.4 sees JSON Pass fall slightly. Value Accuracy moves only $-0.007$ to $+0.033$ across the three models tested.
\item Structural metrics mask value errors. 16 of 21 valid text models score $\geq$96\% on Path Recall, Structure Coverage, and Type Safety. But Value Accuracy drops to 0.693--0.830, and Perfect Response Rate collapses to 0.376--0.526. The scaffolding is correct; the content inside it is not.
\item Model size does not predict structured output quality. Phi-4 (14B) scores 0.798 Value Accuracy, above GPT-5 (0.795) and GPT-5~Mini (0.779). Schematron-8B scores 0.754, outperforming GPT-OSS~20B (0.693) with 2.5$\times$ fewer parameters. This aligns with LLMStructBench~\citep{tenckhoff2026llmstructbench}, which found prompting strategy matters more than model size.
\item Structured hallucinations are harder to catch. When a model hallucinates inside a JSON field, the output looks authoritative: correct format, correct types, plausible value. For instance, on an audio record discussing a product's target market, the ground truth is \texttt{"target\_market\_age": "15 to 35 years"} but a model returns \texttt{"target\_market\_age": "25 to 35"}, changing the lower bound to a value not stated in the meeting transcript. Valid JSON, correct schema, and the error is invisible without field-level verification against the source.
\item Performance varies sharply across modalities. Text best Value Accuracy is 0.830 (GLM-4.7), image best is 0.672 (Gemma-4-31B), and audio best is 0.237 (Gemini-2.5-Flash). Even the image leader reaches only 81\% of the text ceiling, and audio drops to less than one-third. Perfect Response follows the same gradient: 0.526 for text, 0.130 for image, near zero for audio ($\leq 0.9\%$ for all models). JSON compliance remains high across all three settings (most models $\geq$80\%), confirming that schema adherence is not the bottleneck: grounded value extraction is.
\item Rankings shift across modalities. GLM-4.7 leads text, Gemma-4-31B leads images, and Gemini-2.5-Flash leads audio: no single model dominates all three. Concretely: GPT-5.4 ranks 3rd on text but 9th on images; Schematron-8B ranks 19th on text but 10th on images; Gemma-4-31B ranks 11th on text but 1st on images. A text-only benchmark would rank these models identically to their text positions, missing their divergent image capabilities entirely. This demonstrates that multi-source evaluation is not redundant: it reveals capability gaps that single-source leaderboards mask.
\item Interfaze-Beta, a model targeted at structured extraction, reaches Overall 0.855 (within 0.015 of GPT-5.4 at 0.870), with per-modality Value Accuracy of 0.821 on text, 0.559 on image, and 0.205 on audio. On image its Value Accuracy is above Phi-4 (0.557), Qwen3-235B (0.552), Gemma-3-27B (0.552), and Qwen3.5-35B (0.550).
\end{enumerate}

\subsection{Error Taxonomy}

We observe five failure types, ordered by production severity:
\begin{enumerate}[leftmargin=*, nosep]
\item \textbf{Parse failures}: invalid JSON syntax. Six models achieve 0\% parse failures on text; GPT-OSS~20B has 13.2\% (659 of 5,000 records). 19 of 21 valid models fall below 2\%.
\item \textbf{Schema violations}: valid JSON but missing required fields or wrong nesting. GPT-5.4 reaches 99.97\% schema compliance; the median across 21 valid text models is 97.5\%, with Gemma-4-31B (94.5\%), GPT-OSS (85.8\%), and Ministral-3 (90.9\%) falling below 95\%.
\item \textbf{Value errors}: correct structure, wrong content. This is the dominant gap: even models with $>$97\% schema compliance show Value Accuracy scores of 0.693 to 0.830, meaning 17--31\% of leaf values are incorrect despite valid structure.
\item \textbf{Missing paths}: the model omits fields entirely. Path Recall ranges from 0.853 (GPT-OSS) to 0.996 (GPT-5.4), indicating that most models recover the structural skeleton but not all of it.
\item \textbf{Type mismatches}: correct value, wrong JSON type (e.g., string \texttt{"42"} vs.\ integer \texttt{42}). Type Safety ranges from 0.858 to 0.9997, generally tracking schema compliance closely.
\end{enumerate}

\vspace{-10pt}

\section{Limitations and Future Work}
\label{sec:limitations}

Ground truth is human-authored with an LLM cross-check (Gemini family) acting as an automated reviewer. For text, five 100-record review passes suggest a $\sim$3\% residual error rate; image and audio were fully reviewed by hand. Human authorship reduces the LLM-as-generator bias common in synthetic benchmarks, but the schemas still reflect our team's design conventions; future versions should incorporate human-designed schemas from production standards such as FHIR and UBL.

Our evaluation also reflects deliberate design choices. We zero semantic scores when responses fail structural checks, and we evaluate images through validated olmOCR markdown and audio through AMI gold transcripts rather than native inputs. Audio in particular is therefore an upper bound: a Whisper-quality ASR front-end~\citep{radford2023whisper} ($\sim$6--10\% WER on AMI) would introduce transcription errors and likely degrade Value Accuracy further. These choices keep the benchmark production-oriented and isolate structured extraction from vision and ASR quality, but they do not measure end-to-end image or audio performance. Extending the benchmark to native image and audio input is a natural next step.

The metrics remain strict in two ways: exact-match scoring penalizes semantic equivalents such as \texttt{"USA"} vs.\ \texttt{"United States"}, and path flattening treats all arrays as ordered. These choices favor precision, but they can over-penalize semantically correct variants and reordered sets. Future work should add semantic-aware comparison (e.g., STED~\citep{wang2025sted}) and order-sensitive flags for arrays.

Finally, the LLM-judge threshold (60/100) and our hard/soft coverage gates are fixed choices rather than separately ablated variables. Sensitivity analysis, constrained decoding baselines (Outlines, XGrammar), additional modalities such as video and code, and a live leaderboard remain future work.

\section{Conclusion}
\label{sec:conclusion}

We present a multi-source structured output benchmark that tests what production systems actually need: not whether a model can produce valid JSON (nearly all of them can), but whether the values inside that JSON are correct, especially when the values feed downstream systems that cannot detect upstream extraction errors.

Schema compliance is high, but correct field values remain much harder.
Performance drops from text to images and then sharply on audio.
Model rankings shift across modalities instead of staying consistent.
In numbers: best Value Accuracy reaches 83.0\% on text, 67.2\% on images, and 23.7\% on audio, while JSON Pass remains substantially higher than Value Accuracy across all three source domains.
These patterns show that schema compliance alone is an insufficient measure of structured output quality. We release the complete benchmark, evaluation code, and model outputs to enable the community to measure what actually matters for production structured output: whether the values are right.

\section*{Acknowledgments}

We thank the HotpotQA team, the AMI Meeting Corpus team, and the Allen AI olmOCR team for the olmOCR-bench document benchmark, for making their datasets publicly available.


\newpage
\appendix

\section{Dataset Statistics}
\label{app:dataset_stats}
Table~\ref{tab:dataset_stats} provides a detailed breakdown of the benchmark across all three source domains. Table~\ref{tab:image_dist} shows the per-category distribution of the 209 image records.

\begin{table}[!htbp]
\centering
\caption{Image source: per-category distribution in the final 209-record benchmark.}
\label{tab:image_dist}
\small
\begin{tabular}{@{}lr@{}}
\toprule
\textbf{Category} & \textbf{Records} \\
\midrule
Headers/Footers & 67 \\
Multi-Column & 62 \\
Tables & 33 \\
Old Scans & 23 \\
Long/Tiny Text & 11 \\
ArXiv Math & 10 \\
Old Scans Math & 3 \\
\midrule
\textbf{Total} & \textbf{209} \\
\bottomrule
\end{tabular}
\end{table}

\begin{table}[!htbp]
\centering
\caption{Dataset statistics across source domains.}
\label{tab:dataset_stats}
\small
\begin{tabular}{@{}lrrr@{}}
\toprule
\textbf{Statistic} & \textbf{Text} & \textbf{Image} & \textbf{Audio} \\
\midrule
Eval records & 5,000 & 209 & 115 \\
Full benchmark & 25,091 & 209 & 115 \\
Source dataset & HotpotQA & OCR PDFs & AMI Corpus \\
Avg. context (tokens) & 919 & 527 & 7,373 \\
Median schema properties & 4 & 5 & 5 \\
Median required fields & 4 & 5 & 5 \\
\% Medium schemas & 39\% & 12\% & 2\% \\
\% Hard schemas & 61\% & 88\% & 98\% \\
Source categories & 1 & 7 & 1 \\
Best Value Accuracy & 0.830 & 0.672 & 0.237 \\
Models evaluated (valid) & 21 & 21 & 20 \\
Best Perfect Resp.\ Rate & 0.526 & 0.130 & 0.000 \\
\bottomrule
\end{tabular}
\end{table}

\section{Comparison with Prior Benchmarks}
\label{app:comparison}
Table~\ref{tab:benchmark_comparison} compares \ours{} with existing structured output benchmarks across four key axes: source domain coverage, what is evaluated (structure vs.\ values), grounding method, and scale.

\begin{table}[!htbp]
\centering
\caption{Systematic comparison with prior benchmarks across four axes. \checkmark = full support, $\circ$ = partial.}
\label{tab:benchmark_comparison}
\scriptsize
\setlength{\tabcolsep}{2.5pt}
\begin{tabular}{@{}lp{1.4cm}p{1.4cm}p{1.4cm}p{1.4cm}p{1.4cm}p{1.4cm}p{1.4cm}@{}}
\toprule
\textbf{Axis} & \rotatebox{65}{\scriptsize JSONSchema-Bench} & \rotatebox{65}{\scriptsize StructEval} & \rotatebox{65}{\scriptsize DeepJSON-Eval} & \rotatebox{65}{\scriptsize LLMStruct-Bench} & \rotatebox{65}{\scriptsize ExtractBench} & \rotatebox{65}{\scriptsize STED} & \rotatebox{65}{\scriptsize \textbf{SOB (Ours)}} \\
\midrule
\multicolumn{8}{@{}l}{\textit{Source domain coverage}} \\
Text                   & \checkmark & \checkmark & \checkmark & \checkmark &            &\checkmark & \checkmark \\
Image                  &            &            &            &            & \checkmark &           & \checkmark \\
Audio                  &            &            &            &            &            &           & \checkmark \\
\midrule
\multicolumn{8}{@{}l}{\textit{What is evaluated}} \\
Schema compliance      & \checkmark & \checkmark & \checkmark & \checkmark & \checkmark &\checkmark & \checkmark \\
Value exact match      &            &            &            & \checkmark & \checkmark &           & \checkmark \\
Token-level F1         &            &            &            & \checkmark &            &           & \checkmark \\
Per-field grounding    &            &            &            &            & $\circ$    &           & \checkmark \\
Nesting depth stress   &            &            & \checkmark &            &            &           & \checkmark \\
\midrule
\multicolumn{8}{@{}l}{\textit{Evaluation methodology}} \\
Hardened metrics       &            &            &            &            &            &           & \checkmark \\
Consistency scoring    &            &            &            &            &            &\checkmark &            \\
Schema-complexity weighting &       &            & \checkmark &            &            &           & \checkmark \\
\midrule
\multicolumn{8}{@{}l}{\textit{Scale}} \\
\# Records             & 10K schemas & 44 tasks  & 2.1K       & 995        & 12.8K flds & --        & 5,324      \\
\# Models              & 6 engines   & 18        & 8          & 22         & 6          & 6         & 21         \\
\# Sources             & 1           & 1         & 1          & 1          & 1          & 1         & 3          \\
\bottomrule
\end{tabular}
\end{table}

\section{Detailed Metric Definitions}
\label{app:metrics}
This section provides complete mathematical definitions for all seven metrics.

\paragraph{JSON Pass Rate.} Combines three checks into a single binary metric:
\begin{equation}
\texttt{json\_pass} = \mathbb{1}\bigl[\text{parse}(r) \in \{\texttt{dict}, \texttt{list}\}\bigr] \cdot \mathbb{1}\bigl[\texttt{jsonschema.validate}(r, s) = \text{ok}\bigr]
\end{equation}
where $r$ is the candidate response and $s$ is the target schema.

\paragraph{Value Accuracy.} Leaf-level exact match:
\begin{equation}
\texttt{truth} = \frac{|\{p \in G : \mathcal{G}[p] = \mathcal{P}[p]\}|}{|G|}
\end{equation}

\paragraph{Faithfulness Score.} Average token-level F1 across ground-truth leaves. For each leaf $p \in G$:
\begin{equation}
\text{F1}(p) = \frac{2 \cdot |\text{toks}(\mathcal{G}[p]) \cap \text{toks}(\mathcal{P}[p])|}{|\text{toks}(\mathcal{G}[p])| + |\text{toks}(\mathcal{P}[p])|}
\end{equation}
where $\text{toks}(\cdot)$ tokenizes after normalization (lowercase, remove articles/punctuation, collapse whitespace). Missing paths contribute F1 = 0.

\paragraph{Path Recall.} $|O| / |G|$ where $O = G \cap P$.

\paragraph{Structure Coverage.} F1 over path sets:
\begin{equation}
\texttt{struct\_cov} = \frac{2 \cdot \frac{|O|}{|P|} \cdot \frac{|O|}{|G|}}{\frac{|O|}{|P|} + \frac{|O|}{|G|}}
\end{equation}

\paragraph{Type Safety.} For each predicted leaf, check if its JSON type matches the schema-expected type (with array indices wildcarded for lookup):
\begin{equation}
\texttt{type\_safety} = \frac{|\{p \in P : \text{type}(\mathcal{P}[p]) = \text{schema\_type}(p)\}|}{|P|}
\end{equation}

\paragraph{Perfect Response Rate.} Binary exact match after canonical key sorting:
\begin{equation}
\texttt{perfect} = \mathbb{1}\bigl[\text{canon}(\mathcal{P}) = \text{canon}(\mathcal{G})\bigr]
\end{equation}

\subsection{Hardening and Coverage Gate Equations}

The hardening factor gates semantic metrics on structural correctness:
\begin{equation}
h = \texttt{json\_parse} \times \texttt{json\_root} \times \texttt{schema\_compliance}
\end{equation}

Hardened metrics:
\begin{align}
\texttt{truth\_score} &= \text{raw} \times h \times \texttt{coverage\_gate} \\
\texttt{faithfulness} &= \text{raw} \times h \times \texttt{coverage\_gate} \\
\texttt{path\_recall} &= \text{raw} \times h \\
\texttt{struct\_cov} &= \text{raw} \times h
\end{align}

Coverage gates by source domain. In both cases the input is the \emph{raw} (pre-hardening) path-set F1 of the prediction against the ground-truth schema, written $f_1^{\text{raw}}$:
\begin{itemize}[nosep]
\item Text (hard): gate $= \mathbb{1}[f_1^{\text{raw}} \geq 0.95]$
\item Image/Audio (soft): gate $= \min\!\bigl(1,\, (f_1^{\text{raw}}/0.90)^2\bigr)$ if $f_1^{\text{raw}} > 0$, else $0$
\end{itemize}

\section{Example Benchmark Record}
\label{app:example}

\begin{figure}[!htbp]
\centering
\begin{minipage}{0.92\textwidth}
\small
\textbf{Context} (truncated): \textit{Hal Ashby was an American film director. Ciro Ippolito is an Italian film director.}

\vspace{0.5em}
\textbf{Question}: Are Hal Ashby and Ciro Ippolito of the same nationality?

\vspace{0.5em}
\textbf{Schema}: object with required fields \texttt{directors}, \texttt{are\_same\_nationality}, and \texttt{reasoning}. Each director contains \texttt{name} and \texttt{nationality}.

\vspace{0.5em}
\textbf{Ground truth}: \texttt{directors = [Hal Ashby / American, Ciro Ippolito / Italian]}, \texttt{are\_same\_nationality = false}.

\vspace{0.5em}
\textbf{Metadata}: complexity = hard, source = HotpotQA, type = comparison.
\end{minipage}
\caption{A complete benchmark record. The model receives context, question, and schema, and must produce JSON matching the ground truth. This record requires multi-hop reasoning, nested structure, and mixed types.}
\label{fig:example_record}
\end{figure}

\subsection{Image Source Example}

\noindent\fbox{\parbox{0.95\textwidth}{\small
\textbf{Category}: tables \quad \textbf{Source}: olmOCR-bench \quad \textbf{Complexity}: hard (8 properties)
\vspace{0.3em}

\textbf{Context} (truncated): \textit{``District Name: NORTH EAST ISD\ldots Campus Name: STONE OAK EL 2016-17\ldots Total Staff 72.0 100.0\%\ldots Teachers 53.0 73.6\%\ldots''}
\vspace{0.3em}

\textbf{Question}: ``Provide a detailed breakdown of the staff information for STONE OAK EL for the 2016-17 academic year, including staff summary and teacher demographics.''
\vspace{0.3em}

\textbf{Schema}: \texttt{campus\_name} (str), \texttt{total\_students} (int), \texttt{staff\_summary} (array of objects), \texttt{teacher\_ethnicity\_breakdown}, \texttt{students\_per\_teacher}
\vspace{0.3em}

\textbf{Ground Truth} (truncated):\\
{\ttfamily\scriptsize
\{"campus\_name": "STONE OAK EL", "total\_students": 849,\\
\ "staff\_summary": [\{"staff\_type": "Total Staff",\\
\ \ \ "campus\_count\_or\_average": 72.0, "campus\_percent": 100.0\},\\
\ \ \{"staff\_type": "Teachers", "campus\_count\_or\_average": 53.0\}],\\
\ "students\_per\_teacher": 16.0\}}
}}

\vspace{0.5em}

\subsection{Audio Source Example}

\noindent\fbox{\parbox{0.95\textwidth}{\small
\textbf{Meeting}: EN2001a \quad \textbf{Speakers}: 5 \quad \textbf{Utterances}: 1,675 \quad \textbf{Complexity}: hard
\vspace{0.3em}

\textbf{Context} (truncated):\\
{\ttfamily\scriptsize
{[}0:11{]} MEE068: DOES ANYONE WANT TO SEE STEVE'S FEEDBACK\\
{[}0:16{]} MEO069: IS THERE MUCH MORE IN IT THAN HE SAID\\
{[}0:21{]} MEE068: NOT REALLY UM JUST WHAT HE'S TALKING ABOUT}
\vspace{0.3em}

\textbf{Question}: ``What are the key technical decisions regarding data handling, and what action items were assigned?''
\vspace{0.3em}

\textbf{Schema}: \texttt{technical\_decisions} (nested), \texttt{action\_items} (array of objects), \texttt{prototype\_plan}, \texttt{meeting\_logistics}
\vspace{0.3em}

\textbf{Ground Truth} (truncated):\\
{\ttfamily\scriptsize
\{"technical\_decisions": \{"data\_handling": \{\\
\ \ \ "primary\_data\_format": "NITE XML framework",\\
\ \ \ "data\_granularity": "Utterance level"\}\},\\
\ "action\_items": [\{"assignee": "MEO069",\\
\ \ \ "task": "Write baseline for prototype"\}]\}}
}}

\section{Value Errors and Format Mismatches Under Exact-Match Scoring}
\label{app:hallucinations}
Table~\ref{tab:hallucinations} shows real examples where models produced structurally valid JSON that nevertheless failed exact-match value scoring. We separate two phenomena that exact match conflates: \emph{(a) genuine value errors} -- fabrication or information loss that any downstream system would treat as wrong -- and \emph{(b) format/paraphrase mismatches} where the model's output is content-faithful to the source but is penalized because exact match cannot recognize semantic equivalence (\S\ref{sec:limitations}).

\begin{table}[!htbp]
\centering
\caption{Examples that passed schema validation but failed exact-match value scoring. \emph{Top:} genuine value errors (fabrication or information loss). \emph{Bottom:} format/paraphrase mismatches that exact match penalizes despite faithful content.}
\label{tab:hallucinations}
\scriptsize
\begin{tabular}{@{}p{1.2cm}p{3.5cm}p{3.5cm}p{3.5cm}@{}}
\toprule
\textbf{Source} & \textbf{Ground Truth} & \textbf{Model Output} & \textbf{Comment} \\
\midrule
\multicolumn{4}{@{}l}{\textit{(a) Genuine value errors}} \\
\midrule
Text & \texttt{"occupations": ["country music artist", "composer"]} & \texttt{"occupations": ["American country music artist", "composer"]} & Inserted ``American'' not present in source text. Semantically plausible but hallucinated. \\
\addlinespace
Audio & \texttt{"target\_market\_age": "15 to 35 years"} & \texttt{"target\_market\_age": "25 to 35"} & Changed lower bound from 15 to 25. Numeric hallucination from a meeting transcript. \\
\addlinespace
Image & \texttt{"project\_name": "ISITEK Project"} & \texttt{"project\_name": "ISITEK"} & Dropped ``Project'' from the OCR source. Partial extraction, not faithful. \\
\addlinespace
Audio & \texttt{"processing\_speed": "around 30 minutes"} & \texttt{"processing\_speed": "A bit more than real time"} & Dropped the specific ``30 minutes'' figure stated in the meeting. Lost numeric detail. \\
\midrule
\multicolumn{4}{@{}l}{\textit{(b) Format/paraphrase mismatches (content faithful)}} \\
\midrule
Image & \texttt{"isbn": "ISBN: 978-1-118-90210-3"} & \texttt{"isbn": "978-1-118-90210-3"} & Prefix stripped; identifier identical. Penalized by exact match, faithful in content. \\
\addlinespace
Audio & \texttt{"data\_deadline": "December 1st"} & \texttt{"data\_deadline": "First of December"} & Reformatted date. Semantically identical but fails exact match against transcript wording. \\
\addlinespace
Audio & \texttt{"evaluation\_scale": "1 (extremely true) to 7 (not true at all)"} & \texttt{"evaluation\_scale": "Scale of one to seven..."} & Digit-to-word conversion plus truncation. Partly faithful, partly information loss. \\
\bottomrule
\end{tabular}
\end{table}

\section{Worked Scoring Example}
\label{app:scoring_example}
To illustrate how metrics are computed, consider a record where the ground truth is:

{\ttfamily\scriptsize
\{"name": "Hal Ashby", "nationality": "American", "is\_alive": false\}}

\noindent and a model returns:

{\ttfamily\scriptsize
\{"name": "Hal Ashby", "nationality": "United States", "is\_alive": false\}}

\noindent The ground truth has 3 leaf paths. Scoring:

\begin{itemize}[nosep, leftmargin=*]
\item \textbf{JSON Pass Rate}: 1 (valid JSON, structured root, schema validates)
\item \textbf{Value Accuracy}: 2/3 = 0.667 (``name'' and ``is\_alive'' match exactly; ``nationality'' does not: ``American'' $\neq$ ``United States'')
\item \textbf{Faithfulness}: mean of per-leaf token F1. ``Hal Ashby'' $\to$ F1=1.0; ``American'' vs ``United States'' $\to$ F1=0.0 (no token overlap); \texttt{false} vs \texttt{false} $\to$ F1=1.0. Mean = 0.667
\item \textbf{Path Recall}: 3/3 = 1.0 (all GT paths present)
\item \textbf{Structure Coverage}: F1 of path sets = 1.0 (identical path sets)
\item \textbf{Type Safety}: 3/3 = 1.0 (string, string, boolean all correct types)
\item \textbf{Perfect Response}: 0 (not an exact full-object match)
\end{itemize}

\noindent This record scores perfectly on all structural metrics but only 0.667 on Value Accuracy, illustrating the core gap our benchmark measures.

\section{Evaluation Categories}
\label{app:categories}
We group the seven reported metrics plus one parse-only auxiliary sub-check (JSON Parse Success) into five categories that distinguish long-context extraction, schema handling, multi-context linking, output reliability, and exact-match strictness. Each category score is the arithmetic mean of its component metrics for a single record, then averaged across records.

\begin{table}[!htbp]
\centering
\caption{Evaluation categories, component metrics, and what each category tests.}
\label{tab:categories_full}
\small
\begin{tabular}{@{}lp{4cm}p{5cm}@{}}
\toprule
\textbf{Category} & \textbf{Metrics} & \textbf{What It Tests} \\
\midrule
Long Context Extraction & Value Accuracy, Faithfulness, Path Recall & Can the model extract correct values from long, multi-paragraph context? \\
\addlinespace
Complex Schema Handling & JSON Pass Rate, Structure Coverage, Type Safety & Can the model follow deeply nested schemas with correct types? \\
\addlinespace
Multi-Context Linking & Value Accuracy, Faithfulness & Can the model connect facts scattered across multiple passages? \\
\addlinespace
Output Contract Reliability & JSON Parse Success, JSON Pass Rate, Type Safety & Does the model consistently produce usable, schema-valid output? \\
\addlinespace
Strict Precision & Perfect Response Rate & Is the full response exactly correct? \\
\bottomrule
\end{tabular}
\end{table}

\section{Dataset Construction Pipeline}
\label{app:construction}
\begin{figure}[!htbp]
\centering
\includegraphics[width=\columnwidth]{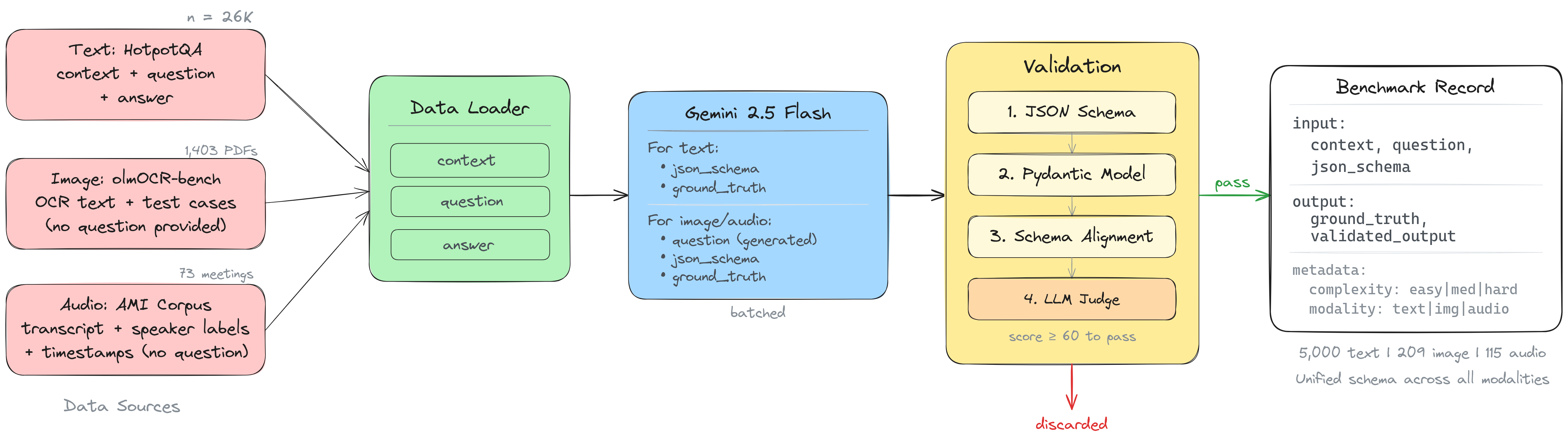}
\caption{Full \ours{} data pipeline. Three source-specific loaders (HotpotQA, olmOCR-bench, AMI Corpus) feed into a human-authoring step, followed by Pydantic validation and an LLM cross-check (Gemini~2.5~Flash for per-record review and Gemini~2.5~Pro for quality scoring) before records are accepted.}
\label{fig:pipeline}
\end{figure}

All three source domains follow identical authoring, validation, and assembly stages, differing only in data loading and question construction. Authored records pass through four checks: JSON Schema validation, Pydantic model validation, schema-to-data alignment, and an LLM cross-check (Gemini~2.5~Pro) scoring accuracy, grounding, and schema quality on a 0--100 scale. Records scoring $<$60 on any dimension are discarded.

\subsection*{Text Source}

The text source draws on HotpotQA~\citep{yang2018hotpotqa}, a multi-hop QA dataset with 112,779 question-answer pairs grounded in Wikipedia passages. Each record provides concatenated supporting paragraphs (distractor split), a question requiring multi-passage reasoning, and a ground-truth answer. For each QA pair, a human reviewer constructs a JSON schema and ground-truth answer that structures the original question, with Gemini~2.5~Flash providing initial drafts and consistency checks that the reviewer accepts or edits. Schemas are designed to capture reasoning structure (entities, relationships, attributes, and the final answer) rather than wrapping the answer in a single field. We enforce a minimum of three schema properties. From this corpus, our authoring-and-filter pipeline produces 25{,}091 \ours{} records, of which \numtext{} comprise the held-out test split used for model evaluation in this paper. Schema complexity distribution: 39\% medium, 61\% hard, with a median of 4 properties per schema.

\subsection*{Image Source}

The image source (209 records) is derived from the olmOCR-bench document benchmark~\citep{poznanski2025olmocr}, which provides 1,403 PDFs with 7,010 deterministic test cases across seven document types: multi-column layouts, dense tables, headers and footers, scanned historical documents, small-print text, mathematical typesetting, and scanned math documents (see Appendix~\ref{app:dataset_stats} for per-category record counts). Each PDF is processed through a mix of robust OCR and human validation, producing a markdown transcript. From the full 1,403 PDF-OCR pairs, we selected 500 candidates with $\geq$50\% quality pass rate, and after human schema authoring and LLM cross-check, 209 records survived. Questions are tailored to document type: table lookups for tabular data, reading-order questions for multi-column layouts, formula extraction for mathematical content, and content comprehension for scanned historical documents. Average context length is $\sim$530 tokens.

\subsection*{Audio Source}

The audio source (115 records) uses the AMI Meeting Corpus via the \textit{edinburghcstr/ami} dataset~\citep{ami_corpus} on HuggingFace, which provides per-utterance gold human transcripts with speaker identifiers and millisecond-precision timestamps. Meetings contain 3--5 speakers with an average of $\sim$800 utterances per meeting and durations from 15 to 85 minutes. We group utterances by meeting, sort by timestamp, and format as readable conversation transcripts with speaker labels (e.g., \texttt{[0:11] MEE068: DOES ANYONE WANT TO SEE STEVE'S FEEDBACK}). This produces substantially longer contexts than the other sources, averaging $\sim$7,300 tokens per record. Questions target speaker turns, action items, discussion topics, and conversational flow. The judge additionally evaluates whether questions require genuine multi-turn understanding rather than trivial single-utterance extraction. Because AMI ships gold transcripts, our audio numbers do not include ASR error and should be read as an upper bound for end-to-end deployment.

\end{document}